\def\year{2021}\relax
\documentclass[letterpaper]{article} % DO NOT CHANGE THIS
\usepackage{aaai21}  % DO NOT CHANGE THIS
\usepackage{times}  % DO NOT CHANGE THIS
\usepackage{helvet} % DO NOT CHANGE THIS
\usepackage{courier}  % DO NOT CHANGE THIS
\usepackage[hyphens]{url}  % DO NOT CHANGE THIS
\usepackage{graphicx} % DO NOT CHANGE THIS
\usepackage{natbib}  % DO NOT CHANGE THIS AND DO NOT ADD ANY OPTIONS TO IT
\usepackage{caption} % DO NOT CHANGE THIS AND DO NOT ADD ANY OPTIONS TO IT
\usepackage{tikz}
\usepackage{etoolbox}
\newcommand*{\lowmark}[1]{\lower.7em \hbox{\tikz\draw (0pt, 0pt)%
    circle (.5em) node {\makebox[0.5em][c]{\small #1}};}}
\robustify{\lowmark}
\usepackage{amsmath}
\usepackage{soul}
\newcommand{\uline}[1]{\ul{#1}}

\title{Contextualized Rewriting for Text Summarization}
\author {
    Guangsheng Bao\textsuperscript{\rm 1,2},
    Yue Zhang\textsuperscript{\rm 1,2}\thanks{Corresponding author.}  \\
}
\affiliations {
    \textsuperscript{1} School of Engineering, Westlake University \\
\textsuperscript{2} Institute of Advanced Technology, Westlake Institute for Advanced Study \\
    \{baoguangsheng, zhangyue\}@westlake.edu.cn
}

\begin{document}
\maketitle

\begin{abstract}
Extractive summarization suffers from irrelevance, redundancy and incoherence. Existing work shows that abstractive rewriting for extractive summaries can improve the conciseness and readability. These rewriting systems consider extracted summaries as the only input, which is relatively focused but can lose important background knowledge. In this paper, we investigate contextualized rewriting, which ingests the entire original document. We formalize contextualized rewriting as a seq2seq problem with group alignments, introducing group tag as a solution to model the alignments,  identifying extracted summaries through content-based addressing. Results show that our approach significantly outperforms non-contextualized rewriting systems without requiring reinforcement learning, achieving strong improvements on ROUGE scores upon multiple extractive summarizers.
\end{abstract}

\section{Introduction}
Extractive text summarization systems \cite{Nallapati2017, Narayan2018, Liu2019} work by identifying salient text segments (typically sentences) from an input document as its summary. They have been shown to outperform abstractive systems \cite{Rush2015,Nallapati2016,Chopra2016} in terms of content selection and faithfulness to the input. However, extractive summarizers exhibit several limitations. First, sentences extracted from the input document tend to contain irrelevant and redundant phrases \cite{Durrett2016, Chen2018, Gehrmann2019}. Second, extracted sentences can be weak in their coherence with regard to discourse relations and cross-sentence anaphora \cite{Dorr2003, Cheng2016}.

To address these issues, a line of work investigates post-editing of extractive summarizer outputs. While grammar tree trimming has been considered for reducing irrelevant content within sentences \cite{Dorr2003}, rule-based methods have also been investigated for reducing redundancy and enhancing coherence \cite{Durrett2016}. With the rise of neural networks, a more recent line of work considers using abstractive models for rewriting extracted outputs sentence by sentence \cite{Chen2018, Bae2019, Wei2019, Xiao2020}. Human evaluation shows that such rewriting systems effectively improve the conciseness and readability. Interestingly, existing rewriters do not improve the ROUGE scores compared with the extractive baselines.

\begin{figure}[t]
    \setlength{\fboxrule}{0.5pt}
    \setlength{\fboxsep}{0.2cm}
    \fbox{\parbox{0.95\linewidth}{
        \textbf{Source Document: }
        thousands of live earthworms have been falling from the sky ... {\it a biology teacher discovered the worms on the surface of the snow while he was skiing in the mountains near bergen at the weekend} ... teacher \textbf{karstein erstad} told norwegian news website ...
        
        \vskip 0.2cm
        \textbf{Gold Summary: }
        teacher karstein erstad found thousands of live worms on top of the snow.

        \vskip 0.2cm
        \textbf{Extractive Summary: }
        {\it a biology teacher discovered the worms on the surface of the snow while he was skiing in the mountains near bergen at the weekend}.

        \vskip 0.2cm
        \textbf{Rewritten Summary: }
        biology teacher \textbf{karstein erstad} discovered the worms on the snow.
    }}
    \caption{Example showing that contextual information can benefit summary rewriting.}
    \label{fig:justex}
\end{figure}

\begin{figure*}[t]
    \setlength{\fboxrule}{0.5pt}
    \setlength{\fboxsep}{0.2cm}
    \fbox{\parbox{0.97\textwidth}{
        \setlength{\baselineskip}{15pt}    
        \textbf{Source Document: }
        \uline{our resident coach and technical expert chris meadows has plenty of experience in the sport and has worked with some of the biggest names in golf.}\lowmark{1} chris has worked with more than 100,000 golfers throughout his career. growing up beside nick faldo, meadows learned that success in golf comes through develping a clear understanding of, and being committed to, your objective. a dedicated coach from an early age, he soon realized his gift was the development of others. \uline{meadows simple and holistic approach to learning has been personally shared with more than 100,000 golfers in a career spanning three decades.}\lowmark{2} \uline{many of his instructional books have become best-sellers, his career recently being recognized by the professional golfers' association when he was made an advanced fellow of the pga.}\lowmark{3} chris has been living golf's resident golf expert since 2003.

        \vskip 0.2cm
        \textbf{Rewritten Summary: }
        \uline{chris meadows has worked with some of golf's big names.}\lowmark{1}
        \uline{he has personally coached more than 100,000 golfers.}\lowmark{2}
        \uline{chris was made an advanced fellow of the pga.}\lowmark{3}
    }}
    \caption{Example of three-step summarization process: selecting, grouping and rewriting.}
    \label{fig:guidsumex}
\end{figure*}

Existing abstractive rewriting systems take extracted summaries as the only input. On the other hand, information from the original document can serve as useful background knowledge for inferring factual details. Take Figure~\ref{fig:justex} for example. A salient summary can be made by extracting the sentence ``{\it a biology teacher...weekend.}'' While a rewriter can simplify the sentence for making a better summary, it cannot provide additional details beyond the sentence unless the document context is also considered. For example, the name of the teacher is not given by the extractive summary, but we can infer that the teacher's name is ``{\it karstein erstad}'' from the context sentences, thereby making the summary more informative.

We propose {\bf contextualized rewriting} by using the full input document as a context for rewriting extractive summary sentences. Rather than encoding only the extractive summary, we use a neural representation model to encode the whole input document, representing extractive summary as a part of the document representation. To inform the rewriter of the current sentence being rewritten, we use content-based addressing \cite{Graves2014}. Specifically, as Figure \ref{fig:guidsumex} shows, a unique group tag is used to index each extracted sentence in the source document, matching an increasing sentence index in the abstractive rewriter as the rewriter generates the output, where the group tags \textcircled{\scriptsize 1} \textcircled{\scriptsize 2} \textcircled{\scriptsize 3} are used to guide the first, second and third rewritten summary sentences, respectively.

We choose the BERT \cite{Devlin2019} base model as the document encoder, building both the extractive summarizer and the abstractive rewriter by following the basic models of \citet{Liu2019}. Our models are evaluated on the CNN/DM dataset \cite{Hermann2015}. 
Results show that the contextualized rewriter gives significantly improved ROUGE \cite{Lin2004} scores compared with a state-of-the-art extractive baseline, outperforming a traditional rewriter baseline by a large margin. In addition, our method gives better compression, lower redundancy and better coherence. The contextualized rewriter achieves strong and consistent improvements on multiple extractive summarizers. To our knowledge, we are the first to report improved ROUGE by rewriting extractive summaries. We release our code at https://github.com/baoguangsheng/ctx-rewriter-for-summ.git.

\section{Related Work}
\label{relwork}

Extractive summarizers have received constant research attention. Early approaches such as TextRank \cite{Mihalcea2004} select sentences based on weighted similarities. Recently, \citet{Nallapati2017} use a neural classifier to choose sentences and a selector to rank them. \citet{Chen2018} use a Pointer Network \cite{Vinyals2015} to extract sentences. \citet{Liu2019} use a linear classifier upon BERT. 
This method gives the current state-of-the-art result in extractive summarization, and we choose it for our baseline.

Rewriting systems manipulate extractive summaries for reducing irrelevance, redundancy and incoherence. \citet{Durrett2016} use compression rules to reduce unimportant content within a sentence and make anaphoricity constraints to improve cross-sentence coherence. \citet{Dorr2003} trim unnecessary phrases in a sentence without hurting grammar correctness by finding the syntactic structures of sentences. In contrast to their work, we consider neural abstractive rewriting, which can solve all the above issues more systematically.

Recently, neural rewriting has attracted much research attention.  \citet{Chen2018} use a seq2seq model with the copy mechanism \cite{See2017} to rewrite extractive summaries sentence by sentence. A reranking post-process is applied to avoid repetition, and the extractive model is also tuned by reinforcement learning with reward signals from each rewritten sentence. \citet{Bae2019} use a similar strategy but with a BERT document encoder and reward signals from the whole summary. \citet{Wei2019} use a binary classifier upon a BERT document encoder to select sentences, and a Transformer decoder \cite{Vaswani2017} with the copy mechanism to generate the summary sentence. \citet{Xiao2020} build a hierarchical representation of the input document. A pointer network and a copy-or-rewrite mechanism are designed to choose sentences for copying or rewriting, followed by a vanilla seq2seq model as the rewriter. The model decisions on sentence selecting, copying and rewriting are tuned by reinforcement learning. 
Compared with these methods, our method is computationally simpler thanks to the freedom from using reinforcement learning and the copy mechanism, as most of the methods above do. In addition, as mentioned earlier, in contrast to these methods, we consider rewriting by including a document-level context, and therefore can potentially improve details and factual faithfulness.

Some hybrid extractive and abstractive summarization models are also in line with our work. \citet{Cheng2016} use a hierarchical encoder for extracting words, constraining a conditioned language model for generating fluent summaries. \citet{Gehrmann2019} consider a bottom-up method, using a neural classifier to select important words from the input document, and informing an abstractive summarizer by restricting the copy source in a pointer-generator network to the selected content. Similar to our work, they use extracted content for guiding the abstractive summary. However, different from their work, which focuses on the word level, we investigate {\it sentence-level} constraints for guiding abstractive {\it rewriting}.

Our method can also be regarded as using group tags to guide the reading context during abstractive summarization \cite{Rush2015,Nallapati2016,See2017}, where group tags are obtained using an extractive summary. Compared with vanilla abstractive summarization, the advantages are three-fold. First, extractive summaries can guide the abstractive summarizer with more salient information. Second, the training difficulty of the abstractive model can be reduced when important contents are marked as inputs. Third, the summarization procedure is made more interpretable by associating a crucial source sentence with each target sentence.

\section{Seq2seq with Group Alignments}
As a key contribution of our method, we model contextualized rewriting as a seq2seq mapping problem with group alignments.
For an input sequence $X$ and an output sequence $Y$, a group set $G$ describes a set of segment-wise alignments between $X$ and $Y$. The mapping problem is defined as finding estimation
\begin{equation}
  \hat{Y} = \arg_Y \max_{Y,G} P(Y,G|X),
\label{eq:seq2seq:def1}
\end{equation}
where
\begin{equation}
  X =\{w_i\}|_{i=1}^{|X|}, 
  Y =\{w_j\}|_{j=1}^{|Y|},
  G =\{G_k\}|_{k=1}^{|G|},
\end{equation}
that $|X|$ denotes the number of elements in $X$, $|Y|$ the number of elements in $Y$, and $|G|$ the number of groups. Each group $G_k$ denotes a pair of text segments, one from $X$ and one from $Y$, which belongs to the same group. Taking Figure \ref{fig:guidsumex} as an example, the first extractive sentence from the document and the first sentence from the summary form a group $G_1$.

The problem can be simplified given the fact that for each group $G_k$, the text segment from $X$ is known, while the corresponding segment from $Y$ is dynamically decided during the generation of $Y$. We thus separate $G$ into two components $G_X$ and $G_Y$, and redefine the mapping problem as
\begin{equation}
  \hat{Y} = \arg_Y \max_{Y,G_Y} P(Y,G_Y|X, G_X),  \\
\label{eq:seq2seq:def2}
\end{equation}
where
\begin{equation}
  G_X = \{g_i= k \text{ if } w_i \in G_k \text{ else } 0 \}|_{i=1}^{|X|}, \\
\label{eq:seq2seq:gx}
\end{equation}
\begin{equation}
  G_Y = \{g_j= k \text{ if } w_j \in G_k \text{ else } 0 \}|_{j=1}^{|Y|}, \\
\label{eq:seq2seq:gy}
\end{equation}
so that for each group $G_k$, a group tag $k$ is assigned, through which the text segment from $X$ in group $G_k$ are linked to the segment from $Y$ in the same group. For the example in Figure \ref{fig:guidsumex}, $G_X=\{1,...,1,0,...,0,2,...,2,3,...,3,0,...,0\}$ and $G_Y=\{1,...,1,2,...,2,3,...,3\}$.

In the encoder-decoder framework, we convert $G_X$ and $G_Y$ into vector representations through a shared embedding table, which is randomly initialized and jointly trained with the encoder and decoder. The vector representations of $G_X$ and $G_Y$ are used to enrich vector representations of $X$ and $Y$, respectively. As a result, all the tokens tagged with $k$ in both $X$ and $Y$ have the same vector component, through which a content-based addressing can be done by the attention mechanism \cite{garg2019}. Here, the group tag serves as a mechanism to constrain the attention from $Y$ to the corresponding part of $X$ during decoding. Unlike approaches which modify a seq2seq model by using rules \cite{Hsu2018, Gehrmann2019}, group tag enables the modification to be flexible and trainable.

\begin{figure*}[t]
    \centering\small
    \includegraphics[scale=0.41]{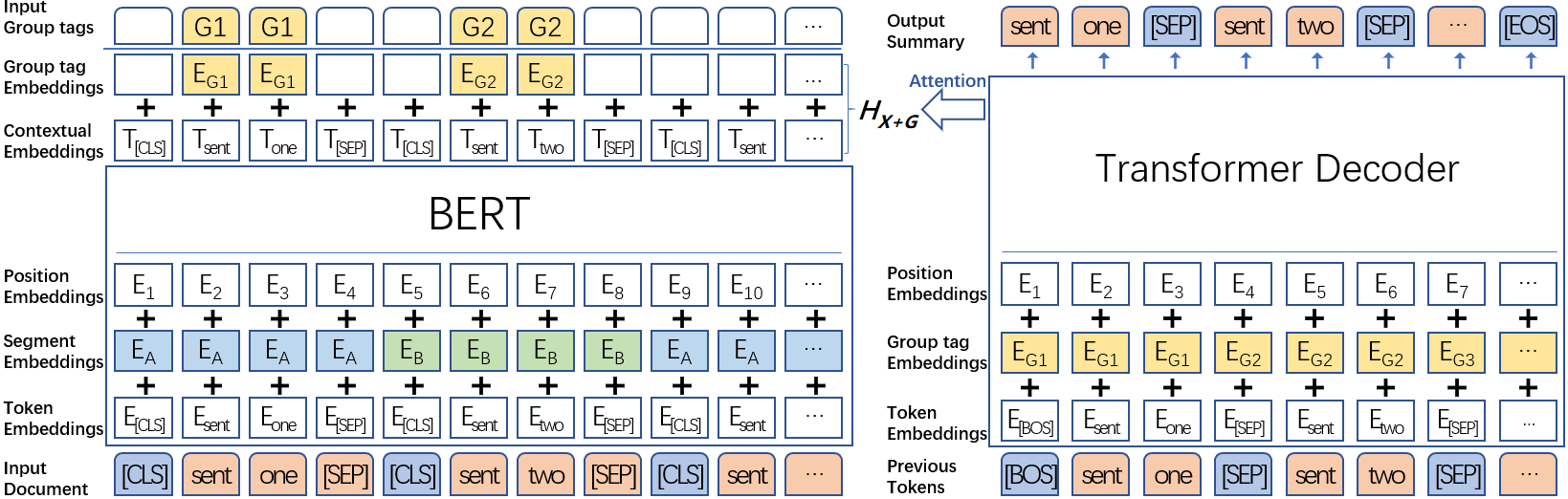}
    \caption{Architecture of the contextualized rewriter. The group tag embeddings are tied between the encoder (left figure) and the decoder (right figure), through which the decoder can address to the corresponding tokens in the document. }
    \label{fig:guidabs}
\end{figure*}

\section{Contextualized Rewriting System}
\label{method}
We take a three-step process in generating a summary. First, an extractive summarization model is used to select a set of sentences from the original document as a guiding source. Second, the guiding source text is matched with the original document, whereby a set of group tags are assigned to each token. Third, an abstractive rewriter is applied to the tagged document, where the group tags serve as a guidance for summary generation.

Formally, we use $X=\{w_i\}|^{|X|}_{i=1}$ to represent document $X$, which contains $|X|$ tokens, and $Y=\{w_j\}|^{|Y|}_{j=1}$ to represent a final resulting summary $Y$, which contains $|Y|$ tokens.

\subsection{Extractive Summarizer}
\label{method:select}
Following \citet{Liu2019}, we use BERT to encode the input document, with a special \textsc{[cls]} token being added to the beginning of each sentence, and interval segments being applied to distinguish successive sentence. On top of the BERT representation of \textsc{[cls]} tokens, an extractor is stacked to select sentences. The extractor uses a Transformer \cite{Vaswani2017} encoder to generate inter-sentence representations, on which, for extracting a summary, an output layer with the sigmoid activation is used to calculate the probability of each sentence being extracted. 

\textbf{Encoder.} 
We use the BERT encoder \textsc{BertEnc} to convert source document $X$ into a sequence of token embeddings $H_X$, taking \textsc{[cls]} embeddings as a representation of the source sentences, denoted as $H_C$.
\begin{equation}
\begin{split}
  H_X &= \textsc{BertEnc}(X) \\
  H_C &= \{H_X^{(i)}|w_i=\textsc{[cls]}\}|^{|X|}_{i=1}.
\end{split}
\end{equation}

\textbf{Extractor.}
We use a Transformer encoder \textsc{TransEnc} to convert sentence embeddings $H_C$ into final inter-sentence representations $H_F$, and calculate the extraction probability on each sentence according to $H_F$.
\begin{equation}
\begin{split}
  H_F &= \textsc{TransEnc}(H_C)  \\
  P(ext_k|X) &= \sigma(W \cdot H_F^{(k)} + b),  \\
\end{split}
\end{equation}
where $ext_k$ means the $k$-th sentence extracted, and $W$ and $b$ are model trainable parameters. 

Given the sequence of extraction probabilities $\{P(ext_k|X)\}|^{C}_{k=1}$, where $C$ denotes the number of sentences in $X$, we make decision on each sentence according to three hyper-parameters: the minimum number of sentences to extract $min\_sel$, the maximum number of sentences to extract $max\_sel$, and a probability $threshold$. In particular, we sort the $C$ sentences in descending order based on $P(ext_k|X)$, where sentences that rank between $0$ and $min\_sel$ are selected by default, while sentences that rank between $min\_sel$ and $max\_sel$ are decided by comparing the probability value with the threshold. Sentences with a probability above $threshold$ are selected. We decide the hyper-parameter values using dev experiments.

Note that our method is slightly different from the extractive model of \citet{Liu2019}, which extracts the 3 most probable sentences as the summary. For the purpose of rewriting with a strong compression, our method allows to extract more sentences as the summary for better recall.

\subsection{Source Group Tagging}
\label{method:grouptag}
We match the extracted summary with the original document for group tagging, taking each sentence in the extracted summary as a group. So that the first summary sentence and the matched sentence forms group one, the second group two, and so on.
Formally, for document $X$ and extractive summary $E$, the $k$-th summary sentence $E_k$ ($k \in [1,...,K]$) is matched to $X$, where every token in $E_k$ is assigned with a group tag $k$. In particular, Eq \ref{eq:seq2seq:gx} is instantiated as
\begin{equation}
    G_X = \{g_i = k \text{ if } w_i \in E_k \text{ else } 0\}|^{|X|}_{i=1},
\label{eq:grouptag:gx}
\end{equation}
where $G_X$ is the sequence of group tags for document $X$.

\subsection{Contextualized Rewriter}
\label{method:abs}
The contextualized rewriter extends the abstractive summarizer of \citet{Liu2019}, which is a standard Transformer sequence to sequence model with BERT as the encoder. As Figure~\ref{fig:guidabs} shows, to integrate group tag guidance, group tag embeddings are added to both the encoder and the decoder. 
% In particular, for the encoder, group tags are used to augment BERT representations of each token. For the decoder, group tags are used to augment input embeddings.
Formally, for an extractive summary $E$, the set of group tags is a closed set of $[1,...,K]$. We use a lookup table $W_G$ to represent the embeddings of the group tags, which is shared by the encoder and the decoder.

{\bf Encoder.}
The original document is processed in the same way as for the extractive model, where a \textsc{[cls]} token is added for each sentence and interval segments are used to distinguish successive sentences. 
After BERT encoding $\textsc{BertEnc}$, the representation of each token is added to the group tag embedding for producing a final representation
\begin{equation}
\begin{split}
  H_{X+G} &= \textsc{BertEnc}(X) + \textsc{Emb}_{W_G}(G_X),  \\
\end{split}
\end{equation}
where $\textsc{Emb}_{W_G}(G_X)$ denotes the retrieved embeddings from the lookup table $W_G$ for group tag sequence $G_X$.

{\bf Decoder.}
Summary sentences are synthesized in a single sequence with special token \textsc{[bos]} at the beginning, \textsc{[sep]} between sentences, and \textsc{[eos]} at the end. The decoder follows a standard Transformer architecture. 

We treat each sentence in the summary as a group. Consequently, the group tag sequence $G_Y$ is fully determined by the summary $Y$. In particular, all the tokens in the $k$-th summary sentence $Y_k (k \in [1,...,K])$ are assigned with a group tag $k$. Therefore, Eq \ref{eq:seq2seq:gy} is instantiated as 
\begin{equation}
    G_Y = \{g_j = k \text{ if } w_j \in Y_k \text{ else } 0\}|^{|Y|}_{j=1}.
\label{eq:grouptag:gy}
\end{equation}
During decoding, the group tag is generated at each beam search step, starting with 1 after the special token \textsc{[bos]} and increasing by 1 after each special token \textsc{[sep]}.

The embedding of group tag $g_j$ is retrieved from the lookup table $W_G$ by $\textsc{Emb}_{W_G}(g_j)$, and added to the token embedding $\textsc{Emb}(w_j)$ and the position embedding. 
\begin{equation}
\begin{split}
  H_{Y+G} &= \textsc{Emb}(Y) + \textsc{Emb}_{W_G}(G_Y)  \\
  P(w_j|w_{<j},X,G_X) & = \textsc{TransDec}(H_{Y+G}^{(<j)}, H_{X+G}), \\
\end{split}
\label{eq:rewrite}
\end{equation}
where $H_{Y+G}$ represents the tagged token embeddings, and $H_{X+G}$ the encoder outputs. The decoder $\textsc{TransDec}$ predicts token probabilities on position $j$ based on the tagged token embeddings before position $j$ and the encoder outputs $H_{X+G}$, consuming $H_{X+G}$ as the memory keys and values for multi-head attention.
Through which the sequence of group tags $G_Y$ is used by the decoder to address the group tags $G_X$ in the encoder, so that the $k$-th rewritten sentence corresponds to the $k$-th extracted sentence in the document. 

\subsection{Training}
\label{method:train}
We train our extractive summarizer and abstractive rewriter separately on a pre-processed dataset labeled with gold-standard extractions.
To generate gold-standard extraction, we match each sentence in human summary to each document sentence, choosing the sentence with the best matching score as the gold extraction for the summary sentence. Specifically, we use the average recall of ROUGE-1/2/L as the scoring function, which follows \citet{Wei2019}.
Differing from existing work \cite{Liu2019}, which aims to find a {\it set} of sentences that maximizes ROUGE matching with the human summary, we find the best match for each summary sentence. As a result, the number of extracted sentences is the same as the number of sentences in the human summary. This strategy is also adopted by \citet{Wei2019} and \citet{Bae2019}.

After matching summary $Y$ to document $X$, we obtain a gold-standard extraction $E=\{E_k\}|_{k=1}^{K}$. 
For training our extractive model, we convert gold-standard extraction $E$ into a label $l_k$ on each sentence $X_k$ in $X$. We set $l_k=1$ if $X_k \in E$, otherwise $l_k=0$.  We train the model with a binary cross-entropy loss function
\begin{equation}
\begin{aligned}
  L_{ext} = &\frac{1}{C} \sum_{k=1}^{C} \big[ - l_k \cdot \log P(ext_k|X)\\
  & - (1 - l_k) \cdot \log (1 - P(ext_k|X)) \big] ,
\end{aligned}
\end{equation}
where $C$ denotes the number of sentences in $X$.

For training our abstractive rewriter, we convert gold-standard extractions $E$ into group tags $G_X$ following Eq \ref{eq:grouptag:gx}, and train the model with a negative log-likelihood loss
\begin{equation}
  L_{wrt} = \frac{1}{|Y|} \sum_{j=1}^{|Y|} - \log P(w_j|w_{<j},X,G_X).
\end{equation}

\section{Experimental Setup}
We evaluate our model on the CNN/Daily Mail dataset \cite{Hermann2015}, which comprises online news articles with several human written highlights (on average 3.75 per article). There are 312,085 samples in total. We use the non-anonymized version and follow the standard splitting of \citet{Hermann2015}, which includes 287,227 samples for training, 13,368 for dev testing, and 11,490 for testing. 
We preprocess the dataset following \citet{See2017} after splitting sentences with the Stanford CoreNLP \cite{Manning2014}. We tokenize sentences into subword tokens, and truncate documents to 512 tokens.

We evaluate our models automatically using ROUGE \cite{Lin2004}, reporting the unigram overlap ROUGE-1 and the bigram overlap ROUGE-2 as metrics for informativeness, and the longest common subsequence ROUGE-L as an indicator of fluency. All scores are calculated using pyrouge. \footnote{https://pypi.org/project/pyrouge/0.1.3/}

\subsection{Extractive Summarizer}
The document encoder is initialized with pre-trained uncased BERT-base, which has 12 transformer layers and the output embedding size is 768. The Transformer extractor is set to 2 layers with an embedding size of 768 and randomly initialized. We use the Adam optimizer \cite{Kingma2015} with $\beta1=0.9$ and $\beta2=0.999$. The encoder and extractor are jointly trained for a total of 50,000 steps with a learning rate schedule \cite{Vaswani2017}
\[
lr = 2e^{-3} \cdot min(step^{-0.5}, step \cdot warmup^{-1.5}),
\]
where $warmup=10,000$. The model is trained with 2 v100 GPUs for about 9 hours.

For inference, we select sentences according to the hyper-parameters $min\_sel=3$, $max\_sel=5$ and $threshold=0.35$, which are chosen by a grid search to find the best average score of ROUGE 1/2/L on the dev dataset.

\subsection{Contextualized Rewriter}
We initialize the document encoder with pre-trained uncased BERT-base model, and initialize the decoder randomly. The Transformer decoder has 6 layers with an embedding size of 768 and tied input-output embeddings \cite{Press2017}. We use the Adam optimizer and default setting $\beta1=0.9$ and $\beta2=0.999$. The model is trained for a total of 240,000 steps, with 20,000 steps for warming-up of the encoder and 10,000 steps for warming-up of the decoder:
\[
lr_{\textsc{Enc}}=2e^{-3} \cdot min(step^{-0.5}, step \cdot warmup_{\textsc{Enc}}^{-1.5})
\]
\[
lr_{\textsc{Dec}}=0.2 \cdot min(step^{-0.5}, step \cdot warmup_{\textsc{Dec}}^{-1.5}).
\]

We use a learning rate of $0.002$ for the encoder, and $0.2$ for the decode, applying dropout with a probability of 0.2, label smoothing \cite{Szegedy2016} with a factor of 0.1, and word dropout \cite{Bowman2016} with a probability of 0.3 on the decoder. We train the model with 2 GPUs on a v100 machine for about 60 hours.

For inference, we constrain the decoding sequence to a minimum length of 50, a maximum length of 200, a length penalty \cite{Wu2016} with $\alpha=0.95$, and a beam size of 5. During beam search, we block the paths on which a repeated trigram is generated, namely Trigram Blocking \cite{Paulus2018}.

\begin{table*}[t]
    \centering
    % \small
    % \setlength{\abovecaptionskip}{6pt}    
    % \setlength{\belowcaptionskip}{-9pt}    
    \begin{tabular}{lccc}
        \hline
        \bf{Method} & \bf{ ROUGE-1 } & \bf{ ROUGE-2 } & \bf{ ROUGE-L} \\
        \hline
        \multicolumn{4}{c}{Extractive}  \\
        \hline
        LEAD-3 \cite{See2017} & 40.34 & 17.70 & 36.57 \\
        BERTSUMEXT \cite{Liu2019} & 43.25 & 20.24 & 39.63 \\
        \hline
        \multicolumn{4}{c}{Abstractive}  \\
        \hline
        BERTSUMABS \cite{Liu2019} & 41.72 & 19.39 & 38.76 \\
        BERTSUMEXTABS \cite{Liu2019} & 42.13 & 19.60 & 39.18 \\
        \hline
        RNN-Ext+Abs+RL \cite{Chen2018} & 40.88 & 17.80 & 38.54 \\
        BERT-Hybrid \cite{Wei2019} & 41.76 & 19.31 & 38.86 \\
        BERT-Ext+Abs+RL \cite{Bae2019} & 41.90 & 19.08 & 39.64 \\
        BERT+Copy/Rewrite+HRL \cite{Xiao2020} & 42.92 & 19.43 & 39.35 \\
        \hline
        \multicolumn{4}{c}{Our Models}  \\
        \hline
        BERT-Ext & 41.04 & 19.56 & 37.66 \\
        Oracle & 46.77 & 26.78 & 43.32 \\
        \hline
        BERT-Abs & 41.70 & 19.06 & 38.71 \\
        BERT-Ext+ContextRewriter & \bf{43.52}$^*$ & \bf{20.57}$^*$ & \bf{40.56}$^*$ \\
        Oracle+ContextRewriter  & 52.57 & 29.71 & 49.69  \\
        \hline
    \end{tabular}
    \caption{Results. The best scores are in bold, and significantly better scores are marked with * ($p<0.001$, t-test). Ext and Abs denotes extractive and abstractive models, respectively, and RL means reinforcement learning.}
    \label{tab:mainresults}
\end{table*}

\section{Results and Analysis}
\label{sec:results}
We compare our models with existing summarization models before analysing the contextualized rewriter.

\subsection{Automatic Evaluation}
The results are shown in Table \ref{tab:mainresults}. The top section consists of extractive models. The middle section contains abstractive models and hybrid systems with a rewriter. The bottom section lists our models. 
In comparison with BERTSUMEXT, our extractive model BERT-Ext gives lower result due to differences in the extraction goal, as discussed earlier.

Compared with the extractive baseline BERT-Ext, our model BERT-Ext+ContextRewriter improves ROUGE-1/2/L by 2.48, 1.01 and 2.90, respectively. This shows the effectiveness of contextualized rewriting. 
To isolate the effect of the rewriter from the extractive summarizer, we also did an experiment using the Oracle extractive summary as the input to our contextualized rewriter, as Oracle+ContextRewriter shows. The gap between our BERT-Ext+ContextRewriter result and the Oracle+ContextRewriter result shows the room for further improvement when the extractive summarizer becomes stronger.
The row BERT-Abs shows the result of the BERT based abstractive summarizer which copies the structure and settings of BERT-Ext-ContextRewriter excluding the components related to group tags in Figure \ref{fig:guidabs}. A contrast between our BERT-Ext+ContextRewriter model and BERT-Abs model shows the usefulness of the extractive summary for guiding abstractive rewriting. 

Compared to the rewriting system BERT-Hybrid, our BERT-Ext-ContextRewriter increases ROUGE-1/2/L by 1.76, 1.26 and 1.7, respectively. It demonstrates the effectiveness of contextualized rewriting compared to non-contextualized rewriting. Although with the help of reinforcement learning, a better result can be achieved for the non-contextualized rewriting system, as the results of BERT-Ext+Abs+RL and BERT+Copy/Rewrite+HRL shows, the complexity of the algorithm is inevitably increased. Compared with the best rewriting system BERT+Copy/Rewrite+HRL, our contextualized rewriter BERT-Ext-ContextRewriter still shows a significant improvement by 0.6, 1.14 and 1.21 on ROUGE 1/2/L, respectively, despite that our model is purely generative without copying tokens from the source document.

Compared with the strong extractive model BERTSUMEXT, BERT-Ext-ContextRewriter gives a better score across three ROUGE metrics with a significant margin for 0.27, 0.33 and 0.93 on ROUGE-1/2/L, respectively. 
Considering the different length of extractive summaries and rewritten summaries, we normalize ROUGE scores following \citet{sun-etal-2019-compare}. The relative improvement of our model after normalization is even larger, that it improves over BERTSUMEXT by $4\%$ relatively (from 1.47 to 1.53) on normalized score, compared to the improvement by $0.6\%$ relatively (from 43.25 to 43.52) on ROUGE-1.
To our knowledge, we are the first to report improved ROUGE scores compared to a state-of-the-art extractive baseline by using abstractive rewriting. Human evaluation is given in the next section.

We did not include BART \cite{Lewis2019} in the table, which reports ROUGE-1/2/L of 44.16, 21.28 and 40.90, respectively. Different pre-training method and data are used by BART as compared to the models in Table \ref{tab:mainresults}. First, we use BERT-base, while BART for summarization uses a large model. Second, models in Table \ref{tab:mainresults} use only the first 512 tokens of the document, while BART uses 1024 tokens.

\begin{table}[t]
    \centering\small
    \begin{tabular}{@{}lcccc@{}}
        \hline
        \bf{Method} & \bf{Faith.} & \bf{Read.} & \bf{Info.} & \bf{Conc.}  \\
        \hline
        RNN-Ext+Abs+RL & 4.71 & 3.62 & 3.22 & 3.35  \\
        BERTSUMEXT & 5.00 & 3.45 & 3.90 & 3.55  \\
        BERTSUMEXTABS & 4.86 & 4.22 & 3.78 & 3.85  \\
        \hline
        BERT-Ext+ContextRewriter & 5.00 & 4.15 & 4.01 & 3.80  \\
        \hline
    \end{tabular}
    \caption{Human evaluation on faithfulness, readability, informativeness, and conciseness.}
    \label{tab:humaneval}
\end{table}

\begin{table*}[t]
    \centering
    % \small
    % \setlength{\abovecaptionskip}{6pt}    
    % \setlength{\belowcaptionskip}{-9pt}    
    \begin{tabular}{lcccc}
        \hline
        \bf{Method} & \bf{ROUGE-1} & \bf{ROUGE-2} & \bf{ROUGE-L} & \bf{Words} \\
        \hline
        Oracle & 46.77 & 26.78 & 43.32 & 112 \\
        \hspace{1mm} + ContextRewriter (ours) & 52.57 (+5.80) & 29.71 (+2.93) & 49.69 (+6.37) & 63 \\
        \hline
        LEAD-3 & 40.34 & 17.70 & 36.57 & 85 \\
        \hspace{1mm} + ContextRewriter (ours) & 41.09 (\textbf{+0.75}) & 18.19 (\textbf{+0.49}) & 38.06 (\textbf{+1.49}) & 55 \\
        \hline
        BERTSUMEXT w/o Tri-Bloc & 42.50 & 19.88 & 38.91 & 80 \\
        \hspace{1mm} + ContextRewriter (ours) & 43.31 (\textbf{+0.81}) & 20.44 (\textbf{+0.56}) & 40.33 (\textbf{+1.42}) & 54 \\
        \hline
        BERT-Ext (ours) & 41.04 & 19.56 & 37.66 & 105 \\
        \hspace{1mm} + ContextRewriter (ours) & 43.52 (\textbf{+2.48}) & 20.57 (\textbf{+1.01}) & 40.56 (\textbf{+2.90}) & 66 \\
        \hline
    \end{tabular}
    \caption{Results of four extractive summarizers applied with contextualized rewriter. Tri-Bloc means Trigram Blocking. }
    \label{tab:extguidabs}
\end{table*}

\begin{figure*}[t]
   \begin{minipage}{0.65\textwidth}
     \centering
     \includegraphics[width=1\linewidth]{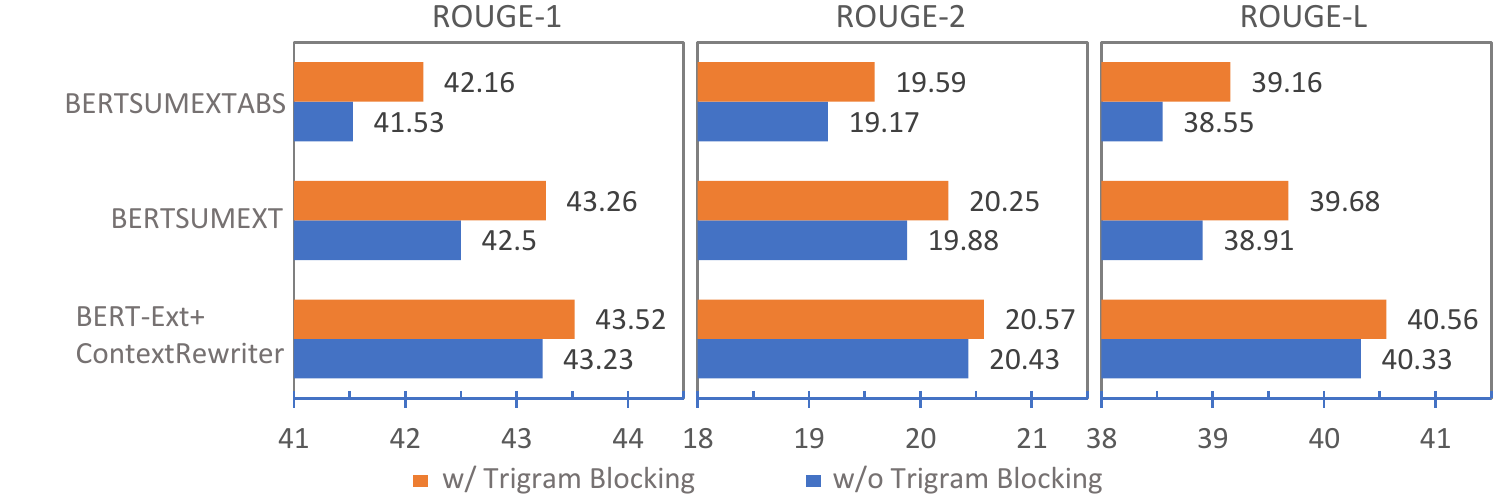}
     \caption{Comparison of the ability to generate non-redundant summaries.}\label{fig:trigramblock}
   \end{minipage}\hfill
   \begin{minipage}{0.3\textwidth}
     \centering
     \includegraphics[width=1\linewidth]{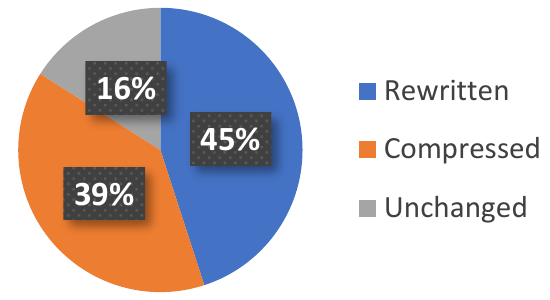}
     \caption{Proportion of rewritten, compressed, and unchanged sentences after rewriting.}\label{fig:compression}
   \end{minipage}
\end{figure*}

\subsection{Human Evaluation}
Intuitively, our model can paraphrase extractive summaries instead of generating summaries from scratch, thereby improving the faithfulness. Furthermore, the abstractiveness of contextualized rewriter can enhance the readability, and the strong compression can improve the conciseness. To confirm these hypothesis, we conduct a human evaluation by randomly select 30 samples from the test set, scoring faithfulness, readability, informativeness, and conciseness from 1(worst) to 5(best) by 3 independent annotators. We report the final result by averaging across annotators.

The result is shown in Table \ref{tab:humaneval}. Compared with non-contextualized rewriter RNN-Ext+Abs+RL, our contextualized rewriter shows obvious advantage across the four aspects. Compared to the extractive baseline BERTSUMEXT, our rewriter enhances the readability, informativeness and conciseness with a significant margin, while keeping the faithfulness. The enhancement of readability is mainly contributed by reduced redundancy and improved coherence. The improvement of conciseness confirms the strong compression of the rewriter. In comparison with the abstractive baseline BERTSUMEXTABS, our rewriter improves faithfulness and informativeness, while keeping the readability and conciseness close. The conciseness of the rewriter is 0.05 lower since it generates summaries for about one word longer than the abstractive model on average. However, by having more text, the rewriter obtains much improved informativeness from 3.78 to 4.01.

\subsection{Universality of the Rewriter}
Our contextualized abstractive rewriter can serve as a general summary rewriter. We evaluate the rewriter with four different extractive summarizers including LEAD-3, BERTSUMEXT, BERT-Ext and Oracle. As Table \ref{tab:extguidabs} shows, the contextualized rewriter improves the summaries generated by all four extractive summarizers. In particular, using LEAD-3 as a basic extractive summarizer, the ROUGE scores improve by a large margin. Even with the best extractive summarizer BERTSUMEXT, the rewriter still enhances the summary quality especially on ROUGE-L, with a 1.42 point improvement. All the extractive summaries are improved by more than 1.4 points on ROUGE-L, which indicates a significant improvement on the fluency. 

In Table \ref{tab:extguidabs}, the ROUGE scores for ``BERTSUMEXT w/o trigram blocking'' is much worse than BERTSUMEXT because there is redundant information. However, when they are applied with our rewriter, they give similar scores where the difference is less than 0.03 point across ROUGE-1/2/L, which is another proof that our rewriter is robust to input of redundant extractive summaries.

\subsection{Analysis}
\label{subsec:analysis}
We further evaluate our rewriter on the ability to reduce irrelevance, improve abstractiveness, and enhance coherence.

\textbf{Redundancy} Redundancy has been a major problem for automatic summarization. Here we study the impact of trigram-blocking to the model performance by comparing with the work of \citet{Liu2019}. As Figure \ref{fig:trigramblock} shows, when the trigram-blocking post-process is removed, all the models give lower ROUGE scores. BERTSUMEXT experiences the most significant drop, while BERTSUMEXTABS has a smaller drop because of less redundancy in an abstractive summarizer. ContextRewriter suffers the least drop, almost halving that by BERTSUMEXTABS, which shows that the contextualized rewriter effectively reduces redundancy. An example shown in Figure \ref{fig:caseofredundancy} demonstrates the ability.

\begin{figure}[t]
    \setlength{\fboxrule}{0.5pt}
    \setlength{\fboxsep}{0.1cm}
    \fbox{\parbox{0.95\linewidth}{
        \setlength{\baselineskip}{15pt}    
        \textbf{Extractive Summary: }
        \uline{oratilwe hlongwane , whose dj name is aj , is still learning to put together words but the toddler is already able to select and play music from a laptop and has become a phenomenon in south africa .}\lowmark{1} 
        \uline{two-year-old oratilwe hlongwane , from johannesburg , south africa , whose dj name is aj , is still learning to put together words but is already able to play music from a laptop , making him a worldwide phenomenon .}\lowmark{2}

        \vskip 0.2cm
        \textbf{Rewritten Summary: }
        \uline{oratilwe hlongwane , whose dj name is aj , is still learning to put together words .} \lowmark{1}
        \uline{he is already able to play music from a laptop , making him a worldwide phenomenon .} \lowmark{2}
    }}
    \caption{Example of the ability to reduce redundancy.}
    \label{fig:caseofredundancy}
\end{figure}

\textbf{Compression} As the column Avg Words in Table \ref{tab:extguidabs} shows, for all the four extractive summarizers, the contextualized rewriter can significantly compress the summaries. For Oracle extractive summaries, it compresses the size by almost a half. For the other models, it compresses the summaries to almost 2/3 of the original summaries on average. 

Looking into the summaries generated by BERT-Ext+ContextRewriter, we find that, as Figure \ref{fig:compression} shows,  $16\%$ of extractive summary sentences are not changed by the rewriter, $39\%$ are compressed, and $45\%$ are rewritten. We obtain these numbers on the test dataset, by adopting the edit-sequence-generation algorithm \cite{Zhang2014} to generate a sequence of word editing actions, mapping an extracted summary sentence to the rewritten one. We categorize a sentence as ``Rewritten'' if the sequence contains an action of adding or modifying, ``Compressed'' if it contains an action of deleting,  and ``Unchanged'' otherwise.

According to 20 samples from the test dataset, all the compressions are on phrases instead of single words. Furthermore, most removed phrases are unimportant, given the fact that only $12\%$ of the removed words are included in reference summaries. For instance, ``{\it they returned to find hargreaves and the girl, who has not been named, lying on top of each other.}'' is compressed into ``{\it they returned to find hargreaves and the girl lying on top of each other.}''

% The average length of reference summaries is 55 words, while that of the output of our model is 66 words. The main difference comes from the length of sentence, where on average the reference is around 3 words shorter than our model output. The average numbers of sentences for reference summary and model output are very similar, which is 3.9 and 3.8 sentences per summary, respectively.

\begin{table}[t]
    \centering\small
    \begin{tabular}{@{}lccc@{}}
        \hline
        \bf{Method} & \bf{1-grams} & \bf{2-grams} & \bf{3-grams} \\
        \hline
        GOLD & 20.66 & 56.55 & 73.48  \\
        BERTSUMEXTABS & 1.39 & 9.81 & 17.79  \\
        BERT-Ext+ContextRewriter & 1.82 & 10.74 & 19.30  \\
        \hline
    \end{tabular}
    \caption{Percentage of novel n-grams.}
    \label{tab:novelngrams}
\end{table}

\textbf{Novel n-grams} As a measure of abstractiveness, we calculate the percentage of novel n-grams as Table \ref{tab:novelngrams} shows the results of 1.82, 10.74 and 19.30. We can see that the contextualized rewriter generates summaries with more novel n-grams compared to BERTSUMEXTABS, which suggests better abstractiveness.

\begin{figure}[t]
    \setlength{\fboxrule}{0.5pt}
    \setlength{\fboxsep}{0.1cm}
    \fbox{\parbox{0.95\linewidth}{
        \setlength{\baselineskip}{15pt}    
        \textbf{Source Document: }
        a university of iowa student has died nearly three months after a fall ... 
        \uline{andrew mogni , 20 , from glen ellyn , illinois , had only just arrived for ...}\lowmark{1}
        \uline{he was flown back to chicago via ...but he died on sunday .}\lowmark{2} ...

        \vskip 0.2cm
        \textbf{Rewritten Summary: }
        \uline{andrew mogni , 20 , from glen ellyn , illinois , had only just arrived for a semester program in italy when the incident happened in january}\lowmark{1}
        \uline{he was flown back to chicago via air ambulance on march 20 , but he died on sunday}\lowmark{2}

        \vskip 0.2cm
        \textbf{Swap Group Tags: }
        \uline{andrew mogni , 20 , was flown back to chicago via air ambulance on march 20 , but he died on sunday}\lowmark{1}
        \uline{he had only just arrived for a semester program in italy when the incident happened in january}\lowmark{2}
    }}
    \caption{Example of the ability to maintain coherence.}
    \label{fig:caseofcoherence}
\end{figure}

\textbf{Coherence} The text generation process of a contextualized rewriter can be controlled by the extractive input, through which we can observe the behavior of the rewriter. Figure \ref{fig:caseofcoherence} uses one output example to demonstrate how the rewriter maintains coherence. We can see that the student name is mentioned in the first summary sentence, while a pronoun is used in the second sentence. As the ``Swap Group Tags'' section shows, when we swap the group tags in the source document, the content of the two summary sentences swap their positions, but the student name is still presented in the first sentence and a pronoun is used in the second sentence. From this case, we can see that the cross-sentence anaphora is maintained correctly.

\section{Conclusion}
We investigate contextualized rewriting of extractive summaries using a neural abstractive rewriter, formalizing the task as a seq2seq problem with group alignments, using group tags to represent alignments, and constraining the attention to rewriting sentence through content-based addressing. Results on standard benchmarks show that using contextual information from the original document is highly beneficial for summary rewriting. Our model outperforms existing abstractive rewriters by a significant margin, achieving strong ROUGE improvements upon multiple extractive summarizers, for the first time. Our method of seq2seq with group alignments is general and can potentially be applied to other NLG tasks.

\section{Acknowledgments}
We would like to thank the anonymous reviewers for their valuable feedback and Wenyu Du for the inspiring discussion. We thank Westlake University High-Performance Computing Center for supporting on GPU resources. This work is supported by the Bright Dream Joint Institute for Intelligent Robotics.

\bibliography{aaai21}

\end{document}